\documentclass{article}

\usepackage{PRIMEarxiv}

\usepackage[utf8]{inputenc} % allow utf-8 input
\usepackage[T1]{fontenc}    % use 8-bit T1 fonts
\usepackage{hyperref}       % hyperlinks
\usepackage{url}            % simple URL typesetting
\usepackage{booktabs}       % professional-quality tables
\usepackage{amsfonts}       % blackboard math symbols
\usepackage{nicefrac}       % compact symbols for 1/2, etc.
\usepackage{microtype}      % microtypography
\usepackage{lipsum}
\usepackage{fancyhdr}       % header
\usepackage{graphicx}       % graphics
\graphicspath{{media/}}     % organize your images and other figures under media/ folder

\title{APP-RUSS: Automated Path Planning for Robotic Ultrasound System}

\author{
  David Liu\\
  Athens Academy\\
  Athens, GA, USA\\
  [3pt] % add vspace for next institution
  Department of Radiology \\
  Massachusetts General Hospital and Harvard Medical School\\
  Boston, MA, USA\\
  \texttt{dliu24@athensacademy.org} \\
   \And
  Jerome Charton, Xiang Li, Quanzheng Li\\
  Department of Radiology \\
  Massachusetts General Hospital and Harvard Medical School\\
  Boston, MA, USA\\
  \texttt{jcharton,xli60,li.quanzheng@mgh.harvard.edu} \\
}

\begin{document}
\maketitle

\begin{abstract}
Autonomous robotic ultrasound System (RUSS) has been extensively studied. However, fully automated ultrasound image acquisition is still challenging, partly due to the lack of study in combining two phases of path planning: guiding the ultrasound probe to the scan target and covering the scan surface or volume. This paper presents a system of Automated Path Planning for RUSS (APP-RUSS). Our focus is on the first phase of automation, which emphasizes directing the ultrasound probe's path toward the target over extended distances. Specifically, our APP-RUSS system consists of a RealSense D405 RGB-D camera that is employed for visual guidance of the UR5e robotic arm and a cubic Bezier curve path planning model that is customized for delivering the probe to the recognized target. APP-RUSS can contribute to understanding the integration of the two phases of path planning in robotic ultrasound imaging, paving the way for its clinical adoption.
\end{abstract}

% keywords can be removed

\section{Introduction} \label{Introduction}
Because of the advantages of being noninvasive, low-cost, portable, and free of radiation, ultrasound has become a major medical imaging modality in healthcare, e.g., for abdomen, cardiovascular, and obstetric imaging \cite{jiang2023robotic,von2021medical,li2021overview}. In parallel, robot-assisted medical imaging enables the controlled trajectory of image acquisition with high precision and accuracy, and it has transformed many significant imaging applications \cite{salcudean2022robot}. As the convergence of these two trends, the Robotic Ultrasound System (RUSS) \cite{graumann2016robotic,suligoj2021robust,priester2013robotic,elek2017robotic,mathiassen2016ultrasound} offers improved reproducibility, enhanced dexterity, and intelligent anatomy and disease-aware imaging \cite{jiang2023robotic}, compared with traditional free-hand ultrasound examinations. In addition, RUSS could substantially relieve the physical strain on sonographers, enable teleoperation that alleviates the concern of the lack of skilled sonographers in rural areas, and satisfy the potential need of separating sonographers from patients to reduce the potential risk of infectious diseases \cite{jiang2023robotic}. Therefore, there has been a dramatic increase in research interest in RUSS and related technologies.    

The operation of RUSS can be typically categorized into teleoperated or autonomous modes. While the teleoperation of RUSS possesses certain advantages, such as reliability and safety, interest in autonomous RUSS has grown more rapidly in recent years. In general, autonomous RUSS could achieve more standardized and reproducible data acquisition, further release sonographers from complex and burdensome ultrasound probe manipulations such as orientation selection, and allow sonographers to focus on medical diagnosis or intervention, which demands professional medical expertise. The key technical obstacles to overcome in developing and deploying clinically plausible autonomous RUSS include a few key steps, including scan target localization and recognition, robotic path planning and execution, robotic arm control, ultrasound probe control and optimization, and anatomy and disease-aware imaging. 

The first step of target recognition has been extensively explored for the guidance and control of robotic ultrasound imaging \cite{ma2021autonomous,ma2021novel,zhang2023visual}, and a well-established OpenCV approach for target recognition \cite{garrido2014automatic} is employed here. Instead, this paper specifically focuses on the second step of path planning and execution for an autonomous RUSS, which can be decomposed into two phases \cite{ma2021autonomous}. In the first phase, the ultrasound scan target is identified, e.g., via computer vision, and a probe landing pose is then estimated and turned into robotic motion \cite{ma2021autonomous}. The second phase further optimizes the ultrasound probe’s placement and orientation to obtain high-quality ultrasound imaging while assuring the patient’s safety \cite{graumann2016robotic,ma2021novel}. While many existing studies of ultrasound probe optimization \cite{jiang2023robotic,von2021medical,li2021overview,graumann2016robotic,ma2021novel} and path planning for scan target coverage \cite{graumann2016robotic,tan2022automatic} have been reported, the path planning for guiding ultrasound probe to the scan target has been much less investigated. Specifically, in earlier RUSS studies using the UR5 robotic arm, e.g., those in \cite{mathiassen2016ultrasound,ma2021novel}, it was assumed that the ultrasound probe is close enough to the scan target, and thus path planning in the first phase of moving the probe to the target was ignored. However, as pointed out in \cite{ma2021autonomous}, it is important to carefully examine the first phase of path planning and seamlessly integrate it with the second phase to enable fully autonomous RUSS in clinical settings, e.g., for ultrasound imaging of liver cancer. Notably, the study in \cite{ma2021autonomous} formulated the first phase of moving the probe to the target as a pose estimation problem instead of performing path planning for a longer distance. However, using probe pose estimation might be insufficient for autonomous RUSS when the ultrasound probe is far from the scan target. Thus, it is crucial to automate the first phase of path planning, which motivates our work in this paper.    

Our system of Automated Path Planning for RUSS, named APP-RUSS, focuses on the first phase of computer vision-based target recognition, performing systematic planning and execution of the ultrasound probe's trajectory towards the target. Specifically, an Intel RealSense D405 RGB-D camera was employed in our APP-RUSS for visual guidance of a UR5e robotic arm, and a cubic Bezier curve path planning method \cite{choi2008path} was adopted and customized in our APP-RUSS for the delivery of the ultrasound probe to the recognized target. The unique innovation and contribution of this work is that our APP-RUSS is fully autonomous, and it can deal with a wide range of scenarios regarding the positions of the ultrasound probe and the scan target, which has been rarely explored yet in the literature, as far as we know. Our work meaningfully adds to the fast-growing field of autonomous RUSS by offering novel insight into the combination of two phases of path planning, that is, guiding the ultrasound probe to the scan target and covering the scan surface or volume in a systematic fashion.

\section{Methods}
\subsection{Hardware and Software Components of APP-RUSS}
\begin{figure}[h!]
  \centering
  \includegraphics[width=0.9\textwidth]{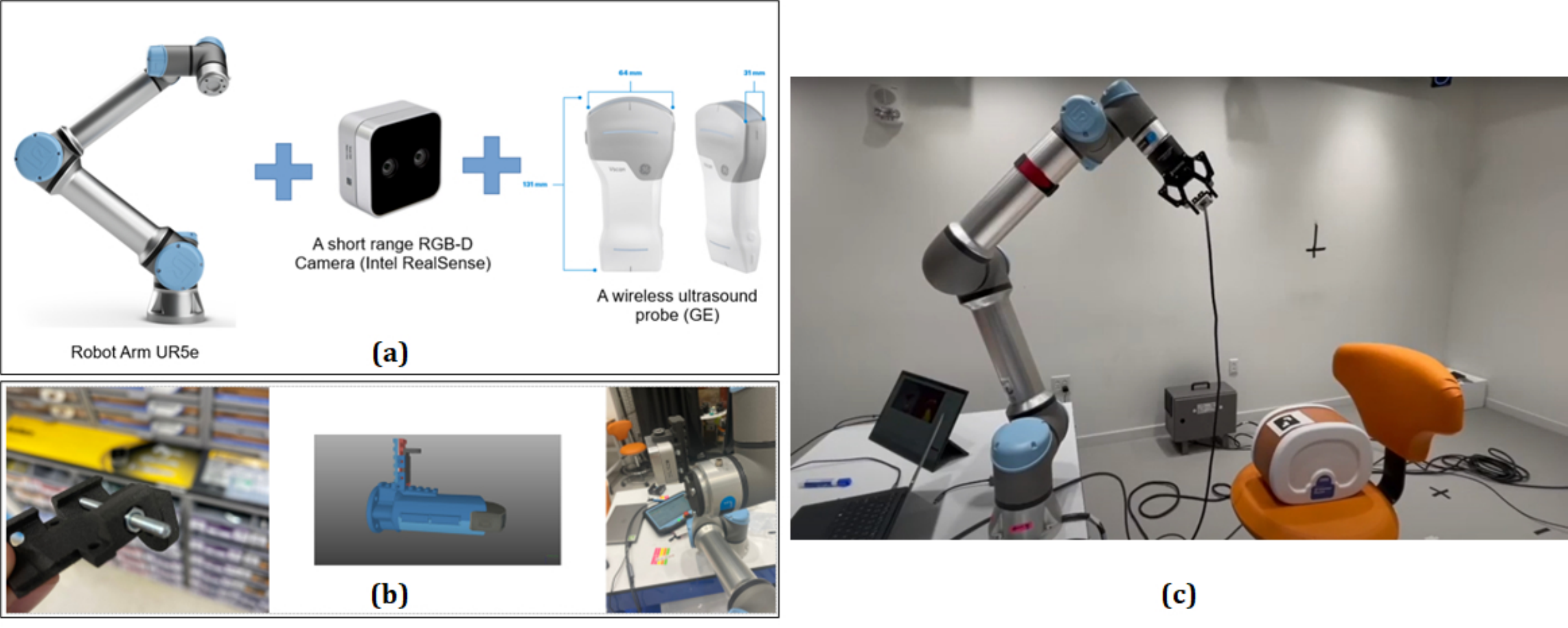}
  \caption{Key components and experimental setup of our APP-RUSS. (a) Hardware components of the robotics system; (b) 3D-printed attachment that mounts the camera and ultrasound probe on the robotic arm; (c) Setting-up of the robotic arm, ultrasound probe, and the abdominal image phantom.}
  \label{fig:fig1}
\end{figure}
APP-RUSS is composed of a UR5e robotic arm (Universal Robotics), an Intel RealSense D405 RGB-D camera, and a wireless ultrasound probe (GE Vscan Air), as shown in Fig. \ref{fig:fig1}(a). Specifically, the UR5e is a light payload industrial collaborative robot that offers sufficient reach (850 mm - 33.5 in) and payload (11 lbs), while being able to perform precise and meticulous tasks such as robotic ultrasound imaging. The RealSense D405 camera has an operating range of 7cm - 50cm, with a depth resolution of 720p and a frames-per-second of 30 fps, which are suitable for our APP-RUSS’ objectives. The GE Vscan Air probe has nice properties of being ultra-portable (dimensions of 131 x 64 x 31 mm, 205 grams), cordless, customizable, and intuitive. The Vscan Air probe can enhance ultrasound imaging with a minimized number of keys and a touchscreen user interface. It supports portrait and landscape modes to optimize image size and ergonomics for different use scenarios. The RealSense camera and Vscan Air probe are mounted onto the UR5e robot via a 3D-printed attachment, as shown in Fig. \ref{fig:fig1}(b). Our APP-RUSS experiment setup is depicted in Fig. \ref{fig:fig1}(c), where an abdominal phantom (CIRS, now part of Sun Nuclear) is used as an ultrasound scan target.

\begin{figure}[h!]
  \centering
  \includegraphics[width=0.6\textwidth]{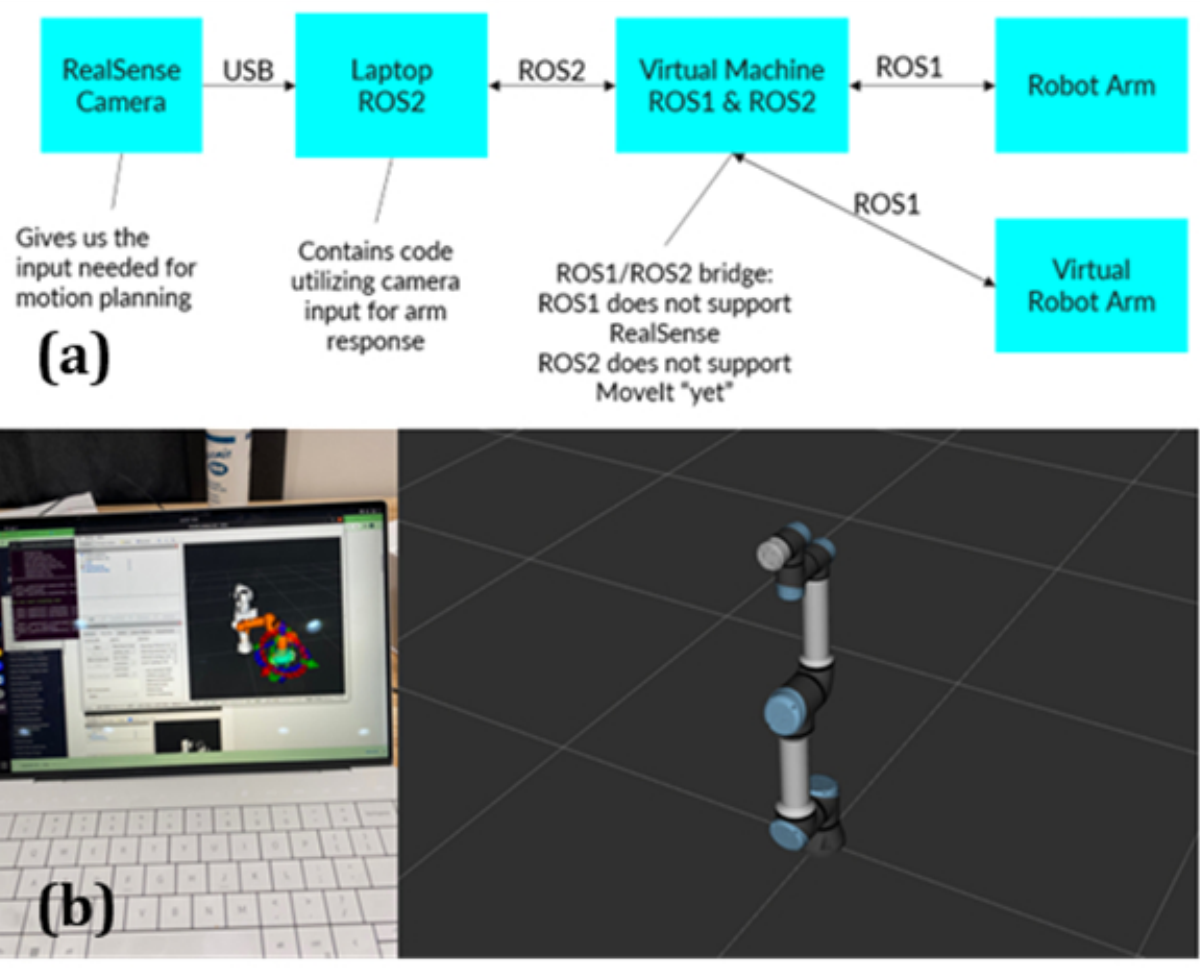}
  \caption{APP-RUSS software framework. (a) Software architecture; (b) Virtual simulation environment. }
  \label{fig:fig2}
\end{figure}
It is nontrivial to integrate the above hardware components and the UR5e robot into a working APP-RUSS system via software platforms and tools. An immediate problem for us to solve was that ROS1 (Robotics Operating System) does not support RealSense D405 camera, and thus ROS2 is needed. However, MoveIt, the virtual motion planning software used for the UR5e, is currently only compatible with ROS1. To solve this software incompatibility problem, we developed a virtual machine on a laptop that bridges both ROS1 and ROS2 and, therefore, integrates the UR5e robot with the RealSense camera, as shown in Fig. \ref{fig:fig2}(a). The laptop virtual machine can also interface with the virtual UR5e robotic arm for simulations. The virtual robotics simulation environment is illustrated in Fig. \ref{fig:fig2}(b), where Gazebo and RViz tools are employed for APP-RUSS’s path planning and execution simulations. These simulations are vital to gain insight and verification of APP-RUSS and its path planning module. The simulations can also generate datasets to train or fine-tune deep learning models that can be transferred to the physical world via zero-shot or few-shot learning \cite{liu2023digital}.

\subsection{Target Recognition and Bezier Curve Path Planning}
\begin{figure}[h!]
  \centering
  \includegraphics[width=0.6\textwidth]{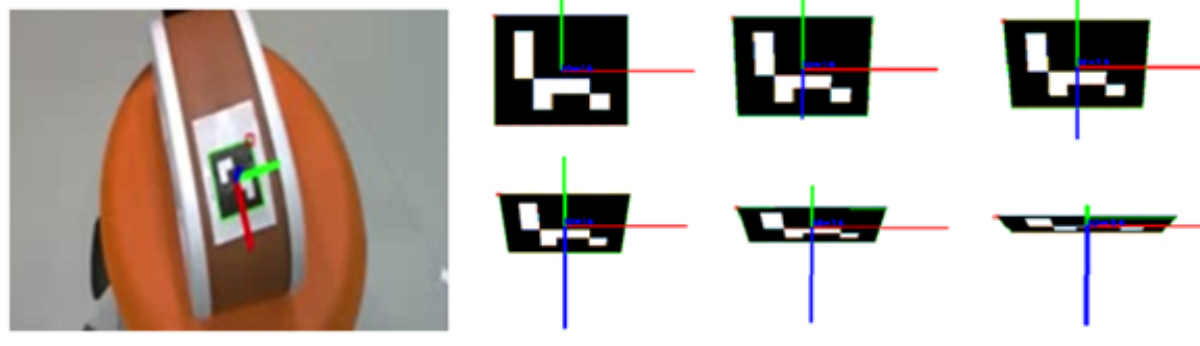}
  \caption{Setup of target recognition with the ArUco marker. }
  \label{fig:fig3}
\end{figure}
As our APP-RUSS is a proof-of-concept study, we employed the commonly used ArUco markers \cite{garrido2014automatic} as the synthesized targets. We used the associated ArUco library based on OpenCV for target recognition and pose estimation \footnote{\url{https://docs.opencv.org/4.x/d5/dae/tutorial_aruco_detection.html}}, as illustrated in Fig. \ref{fig:fig3}. The current ArUco marker recognition module could be replaced by computer vision-based methods, such as the framework in \cite{ma2021novel} where U-Net was used with AC-Kalman tracking to recognize the scan target in real-time, and the model in \cite{huang2018fully} where color and depth channels are used to localize the tissue surface target.

\begin{figure}[h!]
  \centering
  \includegraphics[width=0.9\textwidth]{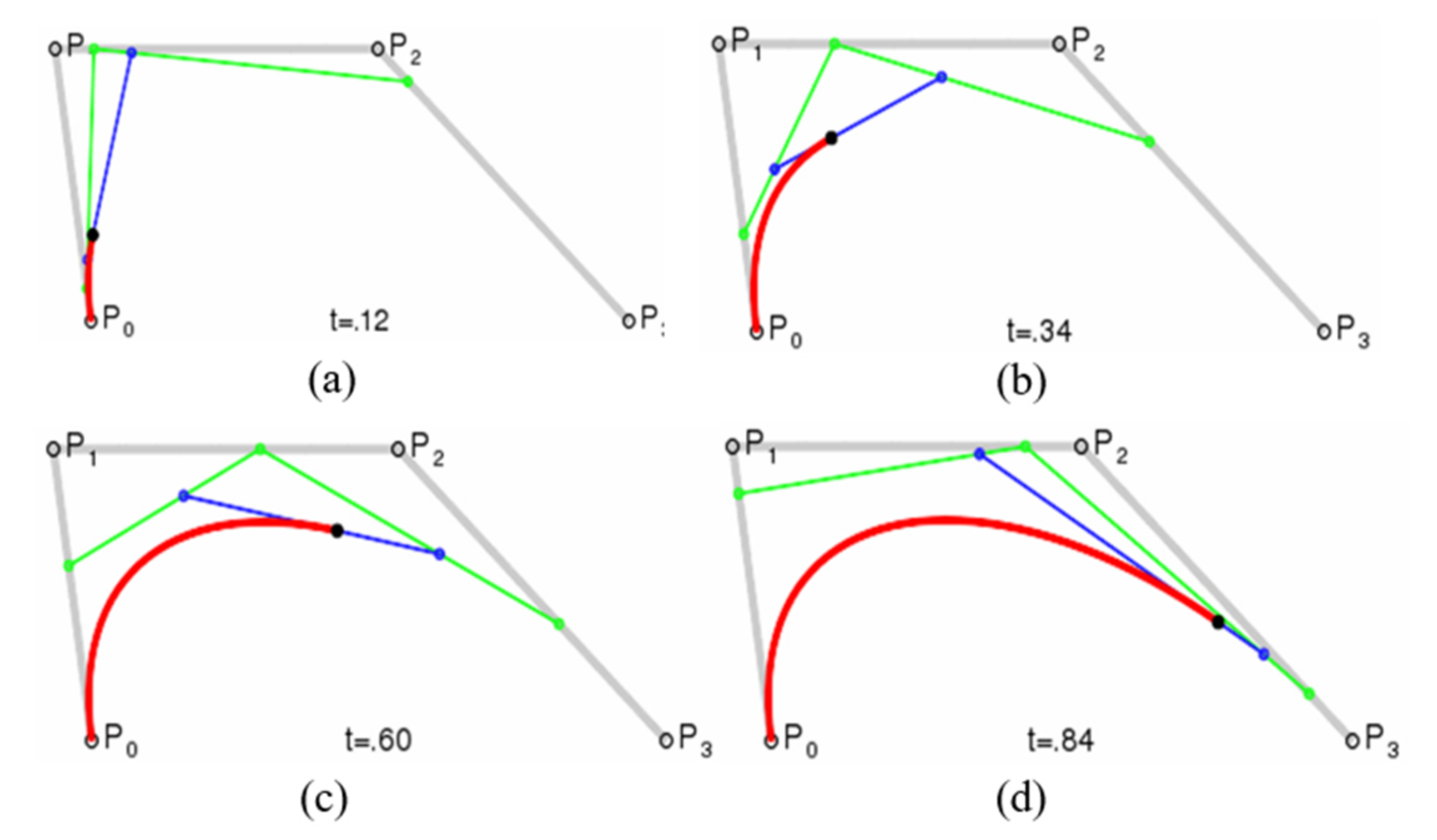}
  \caption{Illustration of the Bezier curve path planning. The red curves in (a) to (d) are the generated path according to the control points $P_0$ to $P_3$.}
  \label{fig:fig4}
\end{figure}

Among the many algorithms for robotic path planning \cite{liu2022hierarchical}, Bezier curves have been widely used for autonomous vehicles using waypoints and corridor constraints \cite{choi2008path}. Because of the Bezier curve’s nice properties in smooth path generation \cite{elhoseny2018bezier} and its ease of computation and stability at the lower degrees of control points, it is adopted and customized in this work for automated path planning in RUSS. The basic idea of Bezier curve path planning is illustrated in Fig. \ref{fig:fig4}, where the path is planned according to the control points ($P_0$-$P_3$) within the bounding polygon. 
\subsection{Experiment Setup}
\begin{figure}[h!]
  \centering
  \includegraphics[width=0.6\textwidth]{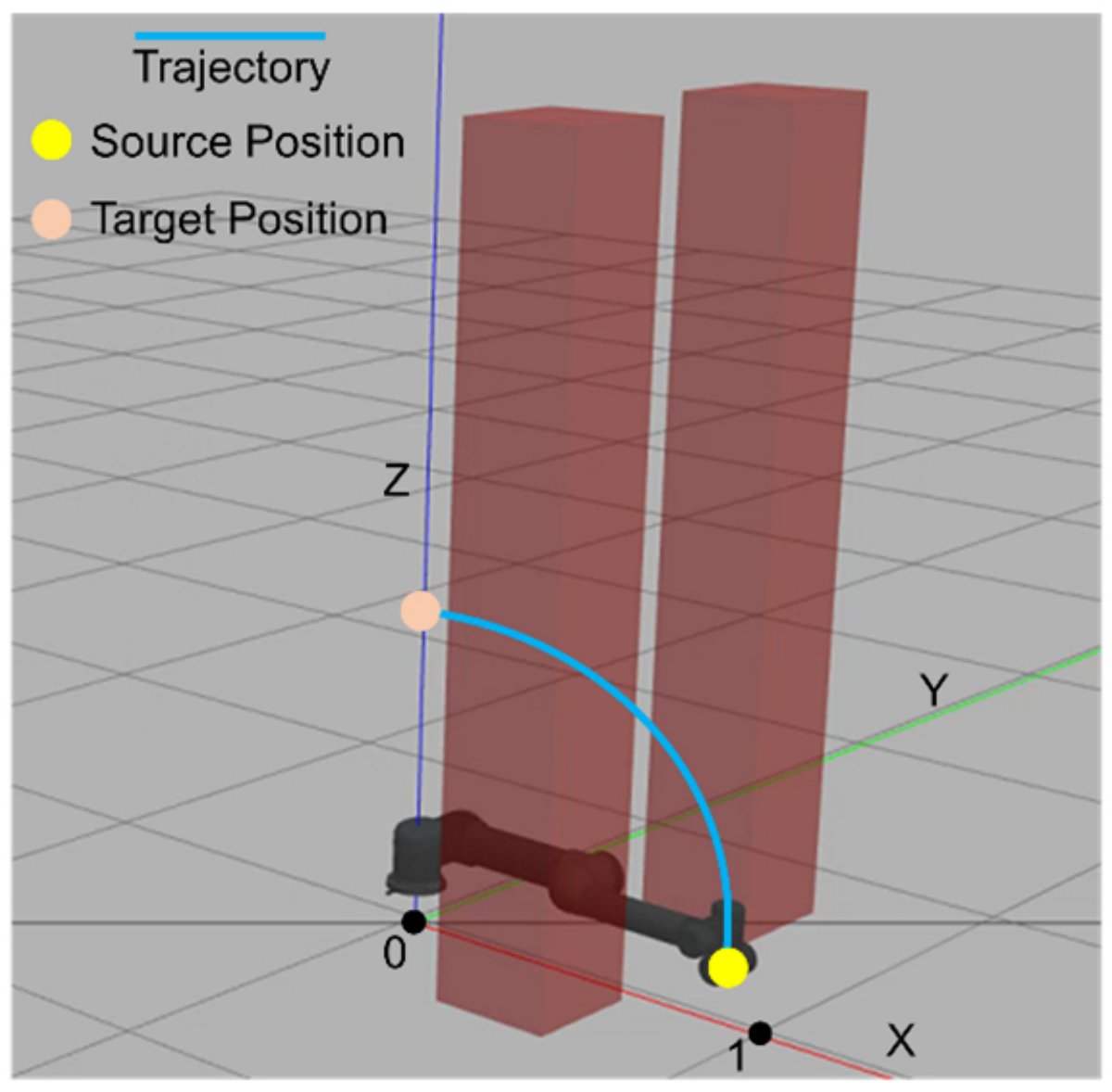}
  \caption{Schematic diagram of the initial position, target position, and obstacle position of the robotic arm in the Gazebo virtual environment. The length of each grid block is 1 unit.}
  \label{fig:fig5}
\end{figure}

In this work, we used MoveIt's UR5e robotic arm model. The simulation of robotic arm control was performed in the Ubuntu 20.04 environment using ROS-Noetic software. In order to test the effectiveness of the Bezier curve-based path-planning model developed in this work, we also set up virtual experiments using Gazebo to simulate real scenarios where the planning of long-distance movements is needed. Specifically, random obstacles are added to the simulation environment. Path planning was then performed according to the position of the source, target, and obstacle. We then executed the simulated robotic movement with and without the path planning. For any given obstacle position, the process was repeated 10 times with random control delays and minor random movements of the robotic arms to simulate the uncertainties in a realistic working condition of the robot. A set of examples for the source, target, and obstacle positions are visualized in Fig. \ref{fig:fig5}. In this figure, the source position is (1,0,0), and the target position is (0,0,1). Two cuboid obstacles were added to the environment to simulate the real working environment of the robotic arm (e.g., equipment around the patients). In this example, the obstacle locations are (0.9, 0.1, 0) and (0.9, -0.1, 0). As the movement of the robotic arm is strictly limited by the position and rotation angle of its joints, the distance between the obstacle and the robotic arm is critical for the movement's success, which can be adjusted in this simulation. Theoretically, the closer the obstacle is to the robotic arm, the more difficult it is to move the robotic arm successfully without touching the obstacle. Based on the recorded movements, we measured the success rate and moving time of the robotic arm controlling in these scenarios with and without path planning. The success rate is defined by the number of times, out of the 10 repetitions, the robotic arm can successfully reach the target point without touching obstacles or stopping due to the limitations in its range of motion. The moving time is defined by the average length of time it takes for the robot arm to move from the source to the target point. We uniformly sample the Bezier curve as the trajectory of the robot arm to simulate its movement along the planned path.

\section{Results and Discussion}
\begin{figure}[h!]
  \centering
  \includegraphics[width=0.6\textwidth]{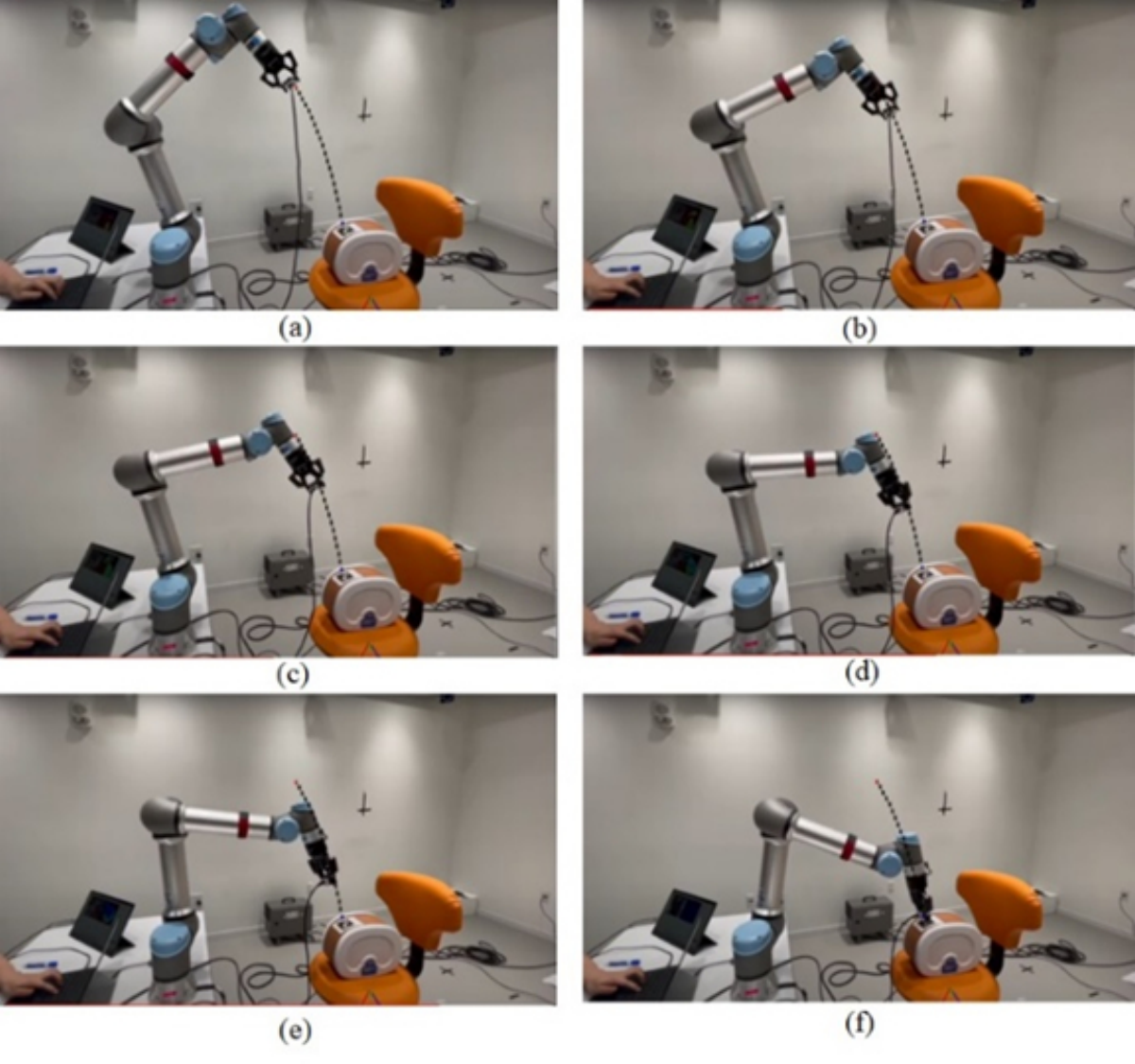}
  \caption{Snapshots of the planned path that guided the ultrasound probe to the ArUco marker target. The black dot curve represents the path. (a)-(f): different time points during the execution of the planned path of the robotic arm.}
  \label{fig:fig6}
\end{figure}

\begin{figure}[h!]
  \centering
  \includegraphics[width=0.6\textwidth]{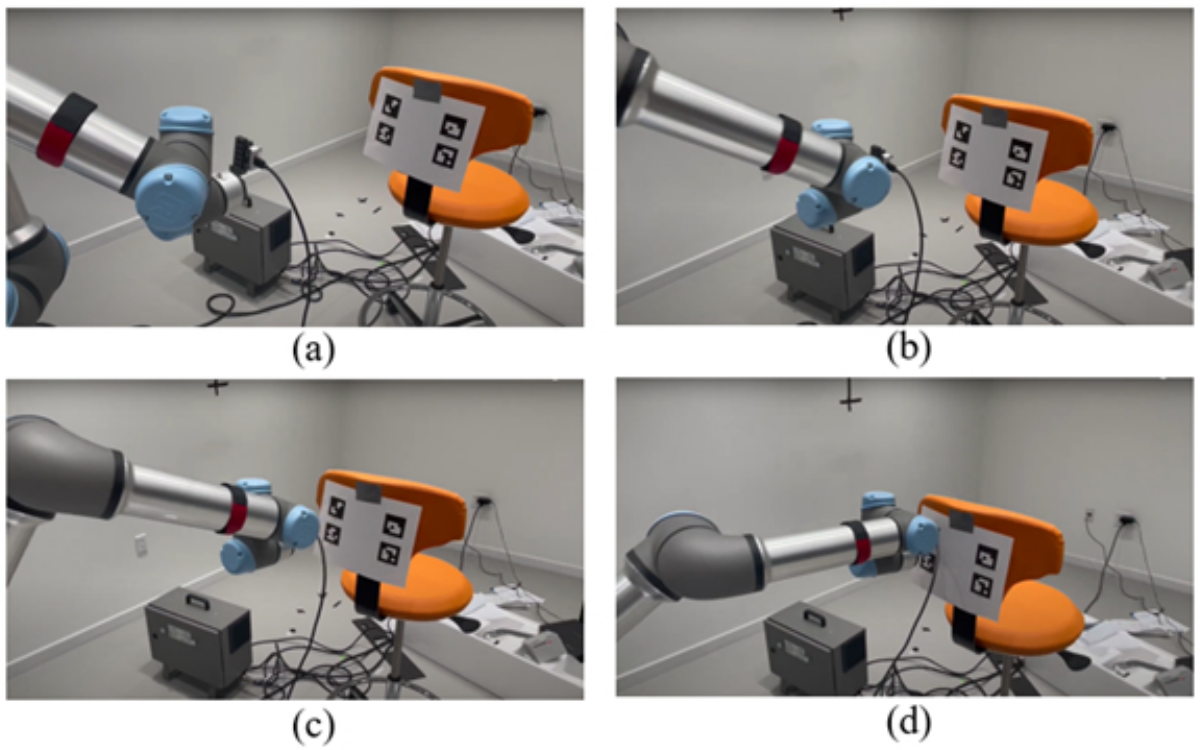}
  \caption{Snapshots of another planned path that guided the ultrasound probe to the ArUco marker target in a different setup. (a)-(d): different points during the robotic execution of the planned path.     }
  \label{fig:fig7}
\end{figure}

We qualitatively evaluated the APP-RUSS prototype in different real-world settings with imaging phantoms and achieved promising results. As demonstrated in Fig. \ref{fig:fig6}, the APP-RUSS can operate in a fully autonomous mode, including recognition of the target, estimation of its pose, calibration of the UR5e robotic arm, planning the path, execution of the robotic movement, and delivery of the ultrasound probe to the target. Notably, the distance between the ultrasound probe and the target in Fig. \ref{fig:fig6} is quite long, which is designed on purpose to evaluate APP-RUSS. Multiple runs of experiments confirmed that APP-RUSS can deal with such long distances of smooth path planning and execution in a fully autonomous fashion. Fig. \ref{fig:fig7} demonstrates a different scenario where multiple ArUco markers are presented. Experimental results confirmed that APP-RUSS could recognize the right target, plan its path accordingly, and execute the path trajectory successfully. The video recordings of the experimental results can be found on YouTube \footnote{\url{www.youtube.com/watch?v=tQlP_7EgDV0}, \url{www.youtube.com/watch?v=I_lrGyHyn1Q}}. 

Based on the random obstacles experiment in the simulation environment introduced in 2.3, we also quantitatively evaluated the effectiveness of path planning of the APP-RUSS model. The performance of the movement execution, both "With Planning" and "Without Planning", are listed in Table \ref{table:1}. Experiment results according to the five different obstacle positions are listed in each corresponding row. The success rate is measured by percentage (number of success movements over 10 repetitions), and the moving time is measured by seconds. From the table, it can be found that regardless of the obstacle position, the success rate is always above 80\% with the path planning model. The moving time is almost always within 40 seconds. While without the path planning model, the success rate and moving time highly depend on the obstacle position. 

\begin{table}
 \caption{Success rate and moving time of the robotic arm movement execution in the simulation experiment with or without path planning. }
  \centering
  \begin{tabular}{lll}
    \toprule
         & With Planning     & Without Planning \\
    \midrule
    No Obstacles & 100\%/29.2s  & 90\%/58.1s     \\
    Obstacle \#1 & 100\%/30.1s  & 60\%/65.7s     \\
    Obstacle \#2 & 90\%/32.5s  & 50\%/64.3s     \\
    Obstacle \#3 & 90\%/32.1s  & 50\%/61.2s     \\
    Obstacle \#4 & 80\%/34.3s  & 30\%/74.9s     \\
    Obstacle \#5 & 90\%/31.3s  & 60\%/68.5s     \\
    No Obstacles & 92\%/31.6s  & 54\%/65.5s     \\
    \bottomrule
  \end{tabular}
\label{table:1}
\end{table}

As mentioned in Section \ref{Introduction}, earlier RUSS studies using the UR5 robotic arm, e.g., those reported in \cite{mathiassen2016ultrasound,ma2021novel}, assumed that the ultrasound probe is close to the scan target. Thus, path planning in the first phase of moving the probe to the target was largely ignored in these studies. Based on the simulation experiment results, it is obvious that path planning over a long distance will be vital for the accuracy and speed of RUSS. Considering the application scenario of RUSS in clinical practice where potential obstacles are abundant, our APP-RUSS work can bridge the gap in the RUSS design by considering a complex environment between the ultrasound probe and the scan target. 

\section{Conclusion}
This paper presented a prototype RUSS system with a path planning model and demonstrated that it can generate and execute a smooth path that guides the ultrasound probe to the target in a complex and long-distance environment. The accuracy of APP-RUSS in guiding the ultrasound probe to the scan target, as measured in the simulation experiment, is quite high. In general, our APP-RUSS work sheds novel insight into the combination of two phases of path planning in robotic ultrasound imaging, contributing to the fast-growing autonomous RUSS field. Our future work will extend the current APP-RUSS in various directions, including automated recognition of scan targets using ultrasound probe and deep learning, path planning for coverage of target organ, optimization of ultrasound probe orientation, and seamless integration of two phases of path planning towards fully autonomous RUSS.

%Bibliography
\bibliographystyle{unsrt}  
\bibliography{references}

\begin{thebibliography}{10}

\bibitem{jiang2023robotic}
Zhongliang Jiang, Septimiu~E Salcudean, and Nassir Navab.
\newblock Robotic ultrasound imaging: State-of-the-art and future perspectives.
\newblock {\em Medical Image Analysis}, page 102878, 2023.

\bibitem{von2021medical}
Felix von Haxthausen, Sven B{\"o}ttger, Daniel Wulff, Jannis Hagenah, Ver{\'o}nica Garc{\'\i}a-V{\'a}zquez, and Svenja Ipsen.
\newblock Medical robotics for ultrasound imaging: current systems and future trends.
\newblock {\em Current robotics reports}, 2:55--71, 2021.

\bibitem{li2021overview}
Keyu Li, Yangxin Xu, and Max Q-H Meng.
\newblock An overview of systems and techniques for autonomous robotic ultrasound acquisitions.
\newblock {\em IEEE Transactions on Medical Robotics and Bionics}, 3(2):510--524, 2021.

\bibitem{salcudean2022robot}
Septimiu~E Salcudean, Hamid Moradi, David~G Black, and Nassir Navab.
\newblock Robot-assisted medical imaging: A review.
\newblock {\em Proceedings of the IEEE}, 110(7):951--967, 2022.

\bibitem{graumann2016robotic}
Christoph Graumann, Bernhard Fuerst, Christoph Hennersperger, Felix Bork, and Nassir Navab.
\newblock Robotic ultrasound trajectory planning for volume of interest coverage.
\newblock In {\em 2016 IEEE international conference on robotics and automation (ICRA)}, pages 736--741. IEEE, 2016.

\bibitem{suligoj2021robust}
Filip Suligoj, Christoff~M Heunis, Jakub Sikorski, and Sarthak Misra.
\newblock Robust--an autonomous robotic ultrasound system for medical imaging.
\newblock {\em IEEE Access}, 9:67456--67465, 2021.

\bibitem{priester2013robotic}
Alan~M Priester, Shyam Natarajan, and Martin~O Culjat.
\newblock Robotic ultrasound systems in medicine.
\newblock {\em IEEE transactions on ultrasonics, ferroelectrics, and frequency control}, 60(3):507--523, 2013.

\bibitem{elek2017robotic}
Ren{\'a}ta Elek, Tam{\'a}s~D Nagy, D{\'e}nes~A Nagy, Bence Tak{\'a}cs, P{\'e}ter Galambos, Imre Rudas, and Tam{\'a}s Haidegger.
\newblock Robotic platforms for ultrasound diagnostics and treatment.
\newblock In {\em 2017 IEEE international conference on systems, man, and cybernetics (SMC)}, pages 1752--1757. IEEE, 2017.

\bibitem{mathiassen2016ultrasound}
Kim Mathiassen, J{\o}rgen~Enger Fjellin, Kyrre Glette, Per~Kristian Hol, and Ole~Jakob Elle.
\newblock An ultrasound robotic system using the commercial robot ur5.
\newblock {\em Frontiers in Robotics and AI}, 3:1, 2016.

\bibitem{ma2021autonomous}
Xihan Ma, Ziming Zhang, and Haichong~K Zhang.
\newblock Autonomous scanning target localization for robotic lung ultrasound imaging.
\newblock In {\em 2021 IEEE/RSJ International Conference on Intelligent Robots and Systems (IROS)}, pages 9467--9474. IEEE, 2021.

\bibitem{ma2021novel}
Guangshen Ma, Siobhan~R Oca, Yifan Zhu, Patrick~J Codd, and Daniel~M Buckland.
\newblock A novel robotic system for ultrasound-guided peripheral vascular localization.
\newblock In {\em 2021 IEEE International Conference on Robotics and Automation (ICRA)}, pages 12321--12327. IEEE, 2021.

\bibitem{zhang2023visual}
Boheng Zhang, Haibo Cong, Yi~Shen, and Mingjian Sun.
\newblock Visual perception and convolutional neural network based robotic autonomous lung ultrasound scanning localization system.
\newblock {\em IEEE Transactions on Ultrasonics, Ferroelectrics, and Frequency Control}, 2023.

\bibitem{garrido2014automatic}
Sergio Garrido-Jurado, Rafael Mu{\~n}oz-Salinas, Francisco~Jos{\'e} Madrid-Cuevas, and Manuel~Jes{\'u}s Mar{\'\i}n-Jim{\'e}nez.
\newblock Automatic generation and detection of highly reliable fiducial markers under occlusion.
\newblock {\em Pattern Recognition}, 47(6):2280--2292, 2014.

\bibitem{tan2022automatic}
Jiyong Tan, Yuanwei Li, Bing Li, Yuquan Leng, Junhua Peng, Jiayi Wu, Baoming Luo, Xinxing Chen, Yiming Rong, and Chenglong Fu.
\newblock Automatic generation of autonomous ultrasound scanning trajectory based on 3-d point cloud.
\newblock {\em IEEE Transactions on Medical Robotics and Bionics}, 4(4):976--990, 2022.

\bibitem{choi2008path}
Ji-wung Choi, Renwick Curry, and Gabriel Elkaim.
\newblock Path planning based on b{\'e}zier curve for autonomous ground vehicles.
\newblock In {\em Advances in Electrical and Electronics Engineering-IAENG Special Edition of the World Congress on Engineering and Computer Science 2008}, pages 158--166. IEEE, 2008.

\bibitem{liu2023digital}
David Liu, Yuzhong Chen, and Zihao Wu.
\newblock Digital twin (dt)-cyclegan: Enabling zero-shot sim-to-real transfer of visual grasping models.
\newblock {\em IEEE Robotics and Automation Letters}, 8(5):2421--2428, 2023.

\bibitem{huang2018fully}
Qinghua Huang, Bowen Wu, Jiulong Lan, and Xuelong Li.
\newblock Fully automatic three-dimensional ultrasound imaging based on conventional b-scan.
\newblock {\em IEEE transactions on biomedical circuits and systems}, 12(2):426--436, 2018.

\bibitem{liu2022hierarchical}
David~W Liu.
\newblock Hierarchical optimal path planning (hopp) for robotic apple harvesting.
\newblock {\em International Journal of Health Sciences and Research}, 2022.

\bibitem{elhoseny2018bezier}
Mohamed Elhoseny, Alaa Tharwat, and Aboul~Ella Hassanien.
\newblock Bezier curve based path planning in a dynamic field using modified genetic algorithm.
\newblock {\em Journal of Computational Science}, 25:339--350, 2018.

\end{thebibliography}

\end{document}